\documentclass{article}

\usepackage{arxiv}
\usepackage{multirow}
\usepackage[utf8]{inputenc} 
\usepackage[T1]{fontenc}    
\usepackage{hyperref}       
\usepackage{url}            
\usepackage{booktabs}       
\usepackage{amsfonts}       
\usepackage{nicefrac}       
\usepackage{microtype}      
\usepackage{lipsum}		
\usepackage{graphicx}
\usepackage{natbib}
\usepackage{doi}

\usepackage{latexsym}
\usepackage{amsmath}
\usepackage{makecell}
\usepackage{subfigure}
\usepackage{wrapfig}

\title{RAP-Net: Region Attention Predictive Network for Precipitation Nowcasting}

\author{ \href{https://orcid.org/0000-0000-0000-0000}{\includegraphics[scale=0.06]{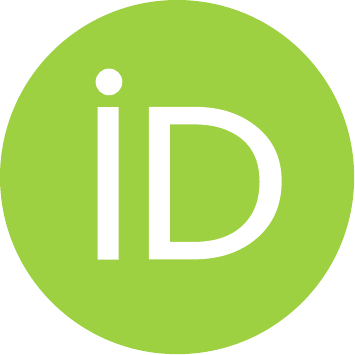}\hspace{1mm}Chuyao Luo}\\
	Department of Computer Science\\
	Harbin Institute Technology, Shenzhen,\\
	\texttt{luochuyao.dalian@gmail.com} \\
	\And
	\hspace{1mm}ZhengZhang \\
	Department of Computer Science\\
	Harbin Institute Technology, Shenzhen,\\
	\texttt{20S151112@stu.hit.edu.cn} \\
}

\renewcommand{\shorttitle}{\textit{arXiv} Template}

\hypersetup{
pdftitle={RAP-Net: Region Attention Predictive Network for Precipitation Nowcasting},
pdfauthor={Chuyao Luo, ZhengZhang, RuiYe, XutaoLi},
pdfkeywords={Spatiotemporal prediction, Precipitation Nowcasting, Neural Networks},
}

\begin{document}
\maketitle

\begin{abstract}
Natural disasters caused by heavy rainfall often cost huge loss of life and property. To avoid it, the task of precipitation nowcasting is imminent. To solve the problem, increasingly deep learning methods are proposed to forecast future radar echo images and then the predicted maps have converted the distribution of rainfall. The prevailing spatiotemporal sequence prediction methods apply ConvRNN structure which combines the Convolution and Recurrent neural network. Although improvements based on ConvRNN achieve remarkable success, these methods ignore capturing both local and global spatial features simultaneously, which degrades the nowcasting in the region of heavy rainfall. To address this issue, we proposed the Region Attention Block (RAB) and embed it into ConvRNN to enhance the forecast in the area with strong rainfall. Besides, the ConvRNN models are hard to memory longer history representations with limited parameters. Considering it, we propose Recall Attention Mechanism (RAM) to improve the prediction. By preserving longer temporal information, RAM contributes to the forecasting, especially in the middle rainfall intensity. The experiments show that the proposed model Region Attention Predictive Network (RAP-Net) has outperformed the state-of-art method. 
\end{abstract}


\section{Introduction}
\label{Introduction}

Precipitation nowcasting has vital influences in the field of transportation, agriculture, tourism industry and city alarming. Due to the higher spatial and temporal resolution of the radar echo image, it is suitable for forecasting the distribution of rainfall by predicting the future radar echo maps and converting each pixel to the rainfall intensity according to the Z-R relationship \cite{shi2017deep}. Therefore, precipitation nowcasting is also defined as the spatiotemporal prediction problem.

Traditional approaches of precipitation nowcasting are motion field-based methods. The specific process can be briefly divided into three parts. First, the motion field is estimated by variational radar echo tracking methods such as optical flow \cite{woo2017operational}. Second, the future radar reflectivities are advected by a semi-Lagrangian advection scheme under the assumption of stationary movement. Third, The performance of forecasts is evaluated by comparing to ground truth. However, these methods do not exploit abundant historical observation.

To overcome the limitation, some deep learning-based methods have been proposed to handle precipitation nowcasting \cite{xingjian2015convolutional,shi2017deep}. It builds a mapping from previous observation to future maps by learning from the abundant historical radar data. Generally, the prevailing approaches utilize the structure of ConvRNN which combines the Convolution Neural Network (CNN) and Recurrent Neural Network (RNN). Furthermore, to enhance the spatiotemporal representation ability,  other types of neural networks such as Spatial Transformer Network (STN) \cite{shi2017deep}, Deformable Convolution Network (DCN) \cite{wu2021motionrnn} and Attention Module \cite{lin2020self} are introduced in the ConvRNN unit and obtain better performance.

However, existing ConvRNN models confront three drawbacks: 1) The convolution employed in the current input only extracts the local features instead of the large-scale representation due to fixed kernel size. It leads that useful information beyond the visual field of convolution dose not be captured and degrades the performance. 2) The convolution applied in the previous hidden state only transmits local previous representation to the current, which causes historical spatial information cannot make fully used. 3) The update process of temporal memory limits the long-term spatiotemporal representation preservation. Thus, the information involving high echo reflectivity is easily dropped. Although some remedial solutions \cite{wang2018eidetic} based on attention mechanism are proposed, they are not suitably applied in the large-scale inputs and long term prediction as to the limitation of the space occupation.

To address the first two problems, we propose the Region Attention Block (RAB) and embed it into the input and hidden state respectively. It exploits the global spatial representation and preserves the local feature simultaneously. RAB classifies each feature map into equal-sized tensors and the similar semantics gathered in the same tensor. Then, the attention module is executed to interact with the contents of all semantics. To this end, the large-scale feature map can be captured from the global view while maintaining local representation. Therefore, the larger spatial feature of the current input and previous hidden state can be preserved. Moreover, to capture the long-term spatiotemporal dependency of representation without increasing uncountable parameters, we present the Recall Attention Mechanism (RAM) to retrieval all historical inputs. More rainfall information is memorized thanks to this component. By adding these modules, the performance for heavy and middle rainfall has significantly improved. In brief, the main contributions of the paper are summarized as follows:

\begin{itemize}
	\item[1.] We first propose a new attention method, name Region Attention Block (RAB), to take capturing both global and local spatial features into account simultaneously to improve the spatial expressivity of feature maps.

	\item[2.] We embed the RAB into the current inputs and previous hidden state to obtain the larger spatial information from the global view and persevere different semantics at the same time. For the same echo with large-scale size and long-range movement between the adjacent time, more useful spatial information can be extracted and predictions in those regions with heavy rainfall are more accurate.  
	
	\item[3.] We propose the Recall Attention Mechanism (RAM) to retrieval all historical inputs without limited parameters. The representation of middle rainfall intensity has maximized extend to be saved in the predicted unit.
\end{itemize}

\section{Related Work}
\label{Related Work}

Traditional methods mainly focus on estimating the motion field between the adjacent radar maps and then the next prediction can be extrapolated based on this movement. Here, the motion field describes the direction and distance of each pixel that need to be moved at the next moment. To obtain the movement,  Tracking Radar Echoes by Correlation (TREC) \cite{wang2013application} divides the whole radar maps into serval equal-sized boxes and calculates the motion vector of each pairs box center by searching the highest correlation between boxes at the adjacent time. Another type of approach is the optical flow-based method \cite{woo2017operational}. It computers the motion field under pixel level based on the assumption that the brightness of pixels remains unchanged. Upon the idea, many algorithms \cite{ryu2020improved} are represented to apply the radar maps with the large movement vector. However, the invariant brightness assumption is a conflict with the realistic movements of hydrometeors and massive historical data are utilized.

To overcome it,  many deep learning-based methods are proposed to predict the radar sequence without the above unreasonable assumption. The prevalent methods apply the structure of ConvRNN. It combines Convolution Neural Network (CNN) and Recurrent Neural Network (RNN) to preserve the spatiotemporal feature of the historical sequence. Furthermore, Wang et.al added a spatial memory in predicted unit \cite{wang2017predrnn,wang2018predrnn++} and attention mechanism in temporal memory \cite{wang2018eidetic} to enhance the spatiotemporal representation ability of short-term and long-term, respectively. Although these methods have remarkable performance, the visualization of predictions is usually burry due to the loss function and the architecture of the model \cite{shouno2020photo}. To handle the issue, Generative Adversarial Network (GAN)  \cite{xie2020energy} has been introduced in the ConvRNN model to improve predictive clarity. Nevertheless, the non-convergence and collapse problem still cause a negative influence on prediction. Moreover, ConvRNN models as the generator still play a decisive role in the framework of GAN.

\section{Preliminary}

The attention function can be described as a mapping from three vectors, name query $Q \in R^{B \times L \times D_q}$, key $K \in R^{B \times L \times D_k}$ and value $V^{B \times L \times D_v}$, to output as the following formula:

\begin{equation}
\label{eq:origianl_attention}
\mbox{Attention(Q,K,V)} = softmax(\frac{f(Q, K^T)}{\sqrt{d_k}}) V,
\end{equation}

\noindent 
where $d_k$ is the dimension of key $K$ and $f$ commonly uses the dot-product(multiplicative).  Attention module is also widely applied in the video prediction \cite{wang2018eidetic,lin2020self}. Specifically, as for a feature map $F_i \in R^{B \times C \times H \times W}$, these three matrixes of query $Q$, key $K$ and value $V$ are generated by three different convolutions and keep the same shape with $F_i$. For spatial attention, the sizes of them are reshaped to $[B \times H*W \times C]$ and feed into the attention function. Here, $f(Q, K^T) \in R^{B \times H*W \times H*W}$ measures the similarity from different positions. For the channel attention, $Q$, $K$ and $V$ convert their sizes to $[B \times C \times H*W]$ and input to attention function. $f(Q, K^T) \in R^{B \times C \times C}$ shows the similarity from different channels. 

\section{Proposed Method}
\label{Proposed Method}

\subsection{Problem Definition}
\label{Problem Definition}
The precipitation nowcasting task can be defined as the spatiotemporal sequence prediction problem \cite{shi2017deep}. Based on historical observations $X_{0:t}$, it aims to forecast the future radar echo images $\bar X_{t+1:T}$ that have maximum probability with ground truth $X_{t+1:T}$ as following:

\begin{equation}
{\bar{X}}_{t+1:T} = \arg\max p({X_{t+1:T}}\big|{X}_{0},{X}_{1}, \cdots, {X}_{t}).
\end{equation}

\noindent In this paper, $t$ and $T$ are set to 5 and 15 respectively, which means that ten continuous radar maps need to be predicted according to five historical images.

\begin{figure}[ht][ht]
	\centering
	\includegraphics[width=0.8\linewidth]{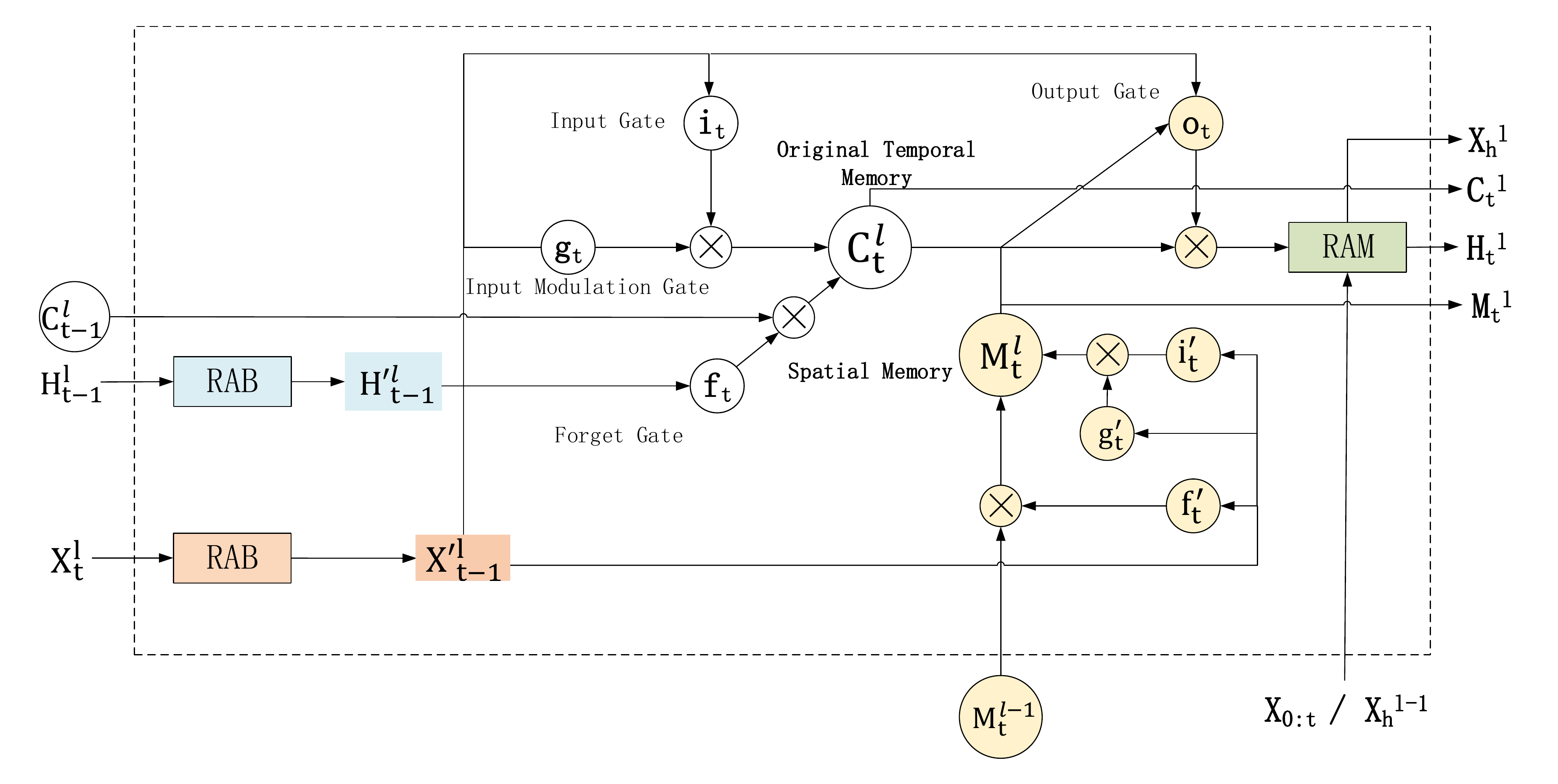}
	\caption{The internal structure of the Region Attention Predictive Unit (RAP-Unit)}
	\label{f1:unit}
\end{figure}

\subsection{Overall Architecture}
\label{Overall Architecture}

The interal structure of Region Attention Unit (RAP-Unit) is shown in Figure~\ref{f1:unit}. The inputs include the current input $X_t^l$, previous hidden state $H_{t-1}^l$, temporal memory $C_{t-1}^l$, spatial memory $M_{t-1}^l$ and long-term historical representation $X_h^{l-1}$, where $X_h^{l-1}$ is all previous inputs $X_{0:t}$ at the bottom layer.  The outputs of RAP-Unit are the current hidden state $H_t^l$, spatial memory $M_t^l$, temporal memory $C_t^l$ and new long-term representation $X_h^l$. The details of calculation are presented according to following formulas:

\begin{small}
	\begin{equation}
	\begin{aligned}
	X_t &= RAB(X_t), \\
	H_{t-1} &= RAB(H_{t-1}), \\
	i_t &= \sigma (W_{xi}*X_t + W_{hi}*H^l_{t-1}  + b_i), \\
	g_t &= tanh(W_{xg}*X_t + W_{hg}*H^l_{t-1} + b_g),\\
	f_t &= \sigma (W_{xf}*X_t + W_{hf}*H^l_{t-1} + b_f ),\\
	i'_t &= \sigma (W'_{xi}*X_{t} + W_{mi}*M^{l-1}_t+ b'_i), \\
	g'_t &= tanh(W'_{xg}*X_{t} + W_{mg}*M^{l-1}_t + b'_g), \\
	f'_t &= \sigma (W'_{xf}*X_{t} + W_{mf}*M^{l-1}_t+b'_f), \\
	C^l_t &= i_t \circ g_t + f_t \circ C^l_{t-1},\\
	M^l_t &= i'_t \circ g'_t + f'_t \circ M^{l-1}_{t},\\
	o_t &= \sigma(W_{xo}*X_t + W_{ho} * H^l_{t-1} + W_{co}*C^l_t + W_{mo}*M^l_t+b_o),\\
	H^l_t &= o_t \circ tanh(W_{1 \times 1}*[X^l_t,M^k_t]),\\
	H^l_t, X_h^l&= RAM(H^l_t,X_h^{l-1}*W_l),
	\end{aligned}
	\label{eq:unit}
	\end{equation}
\end{small}

\noindent where `$*$' and `$\circ$' denote the convolution and Hadamard product respectively. `$i_t$',`$g_t$',`$f_t$',`$i'_t$',`$g'_t$',`$f'_t$' indicate the various gates and are viewed as the intermediate variables. Here, RAB and RAM are the Region Attention Block and Recall Attention Mechanism, respectively. Embedding them can improve the prediction substantially.

\begin{figure}[ht]
	\centering
	\includegraphics[width=1\linewidth]{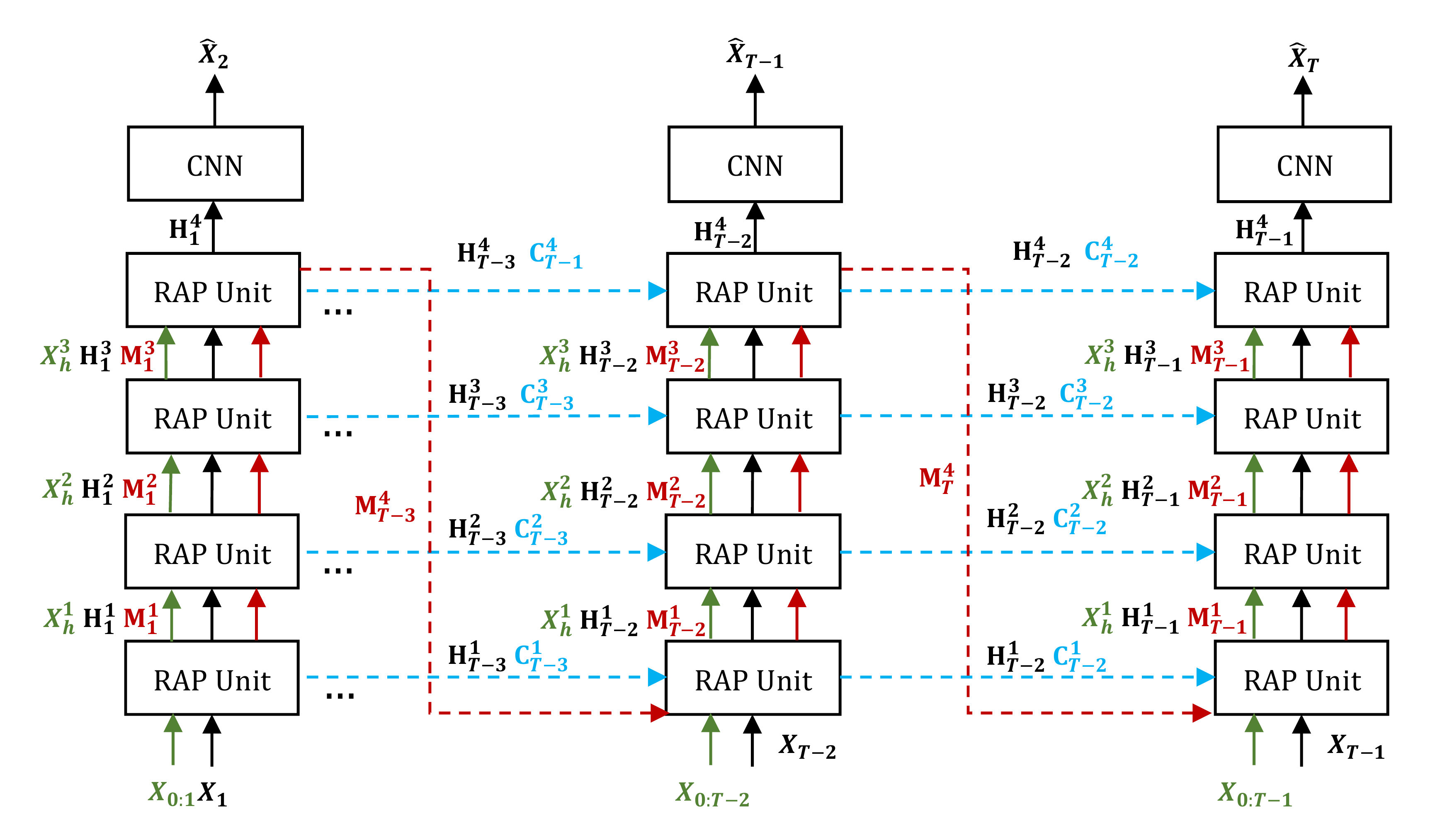}
	\caption{The overall architecture of the Region Attention Predictive Network (RAP-Net)}
	\label{f2:architecture}
\end{figure}

The overall architecture of the proposed model RAP-Net is represented as Figure~\ref{f2:architecture} shown. It refers the structure of PredRNN \cite{wang2017predrnn} and generates the predictions from timestamp $2$ to $T$. The red and blue arrows denote the delivering direction of spatial and temporal memory respectively. Different from the PredRNN, RAP-Net has another data flow to transmit long-term spatiotemporal information $X_h^l$. Besides, we notice that the majority of ConvRNN models employ this architecture. Therefore, the experimental result can be reflected the effectiveness of the predicted unit better.

\begin{figure*}[]
	\centering
	\subfigure[The Similarity Matrix in terms of tradictional attention mechanism.]{
		\label{f3:orgianl_attention} 
		\includegraphics[width=0.45\linewidth]{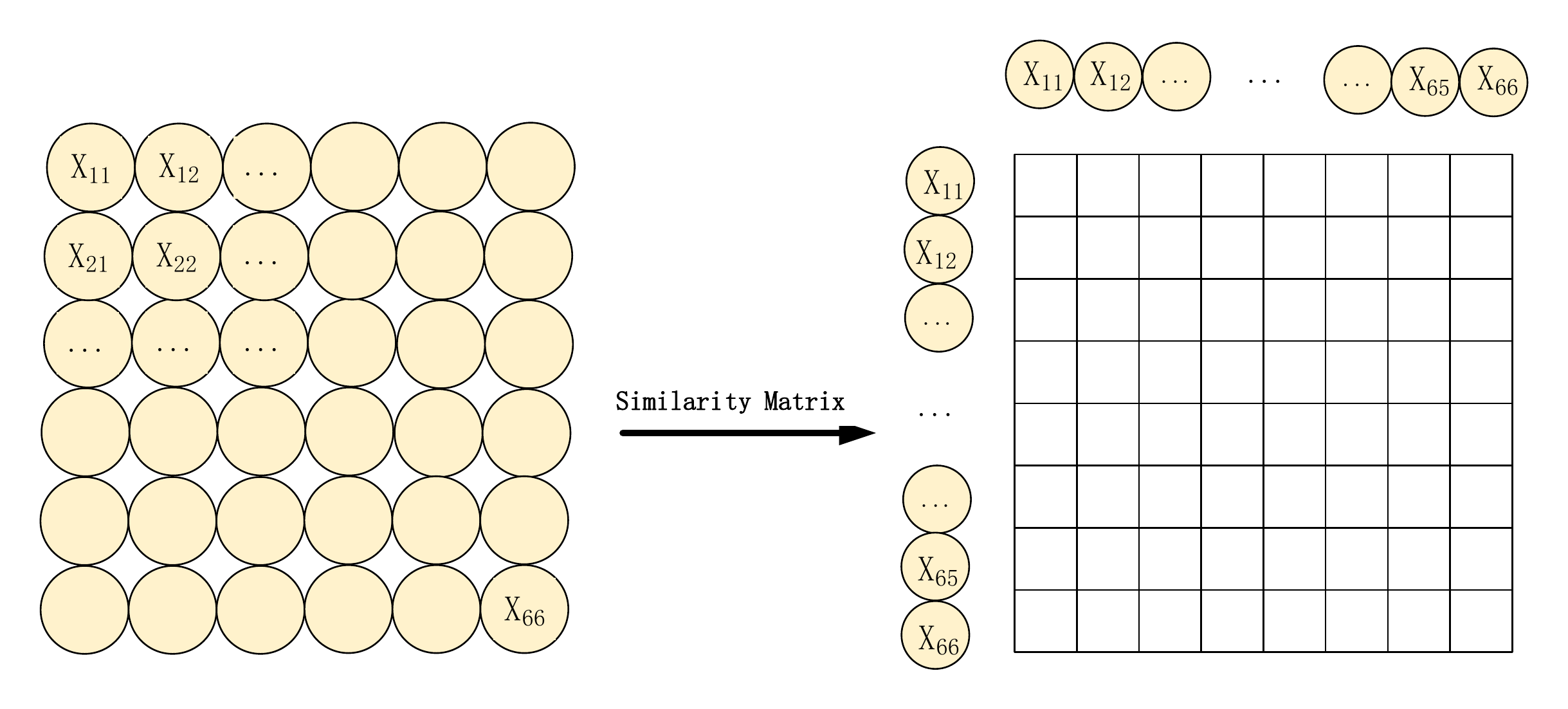}}
	\hspace{0.5in}
	\subfigure[The Similarity Matrix in terms of Vision Transformer.]{
		\label{f4:vit_atetntion} 
		\includegraphics[width=0.45\linewidth]{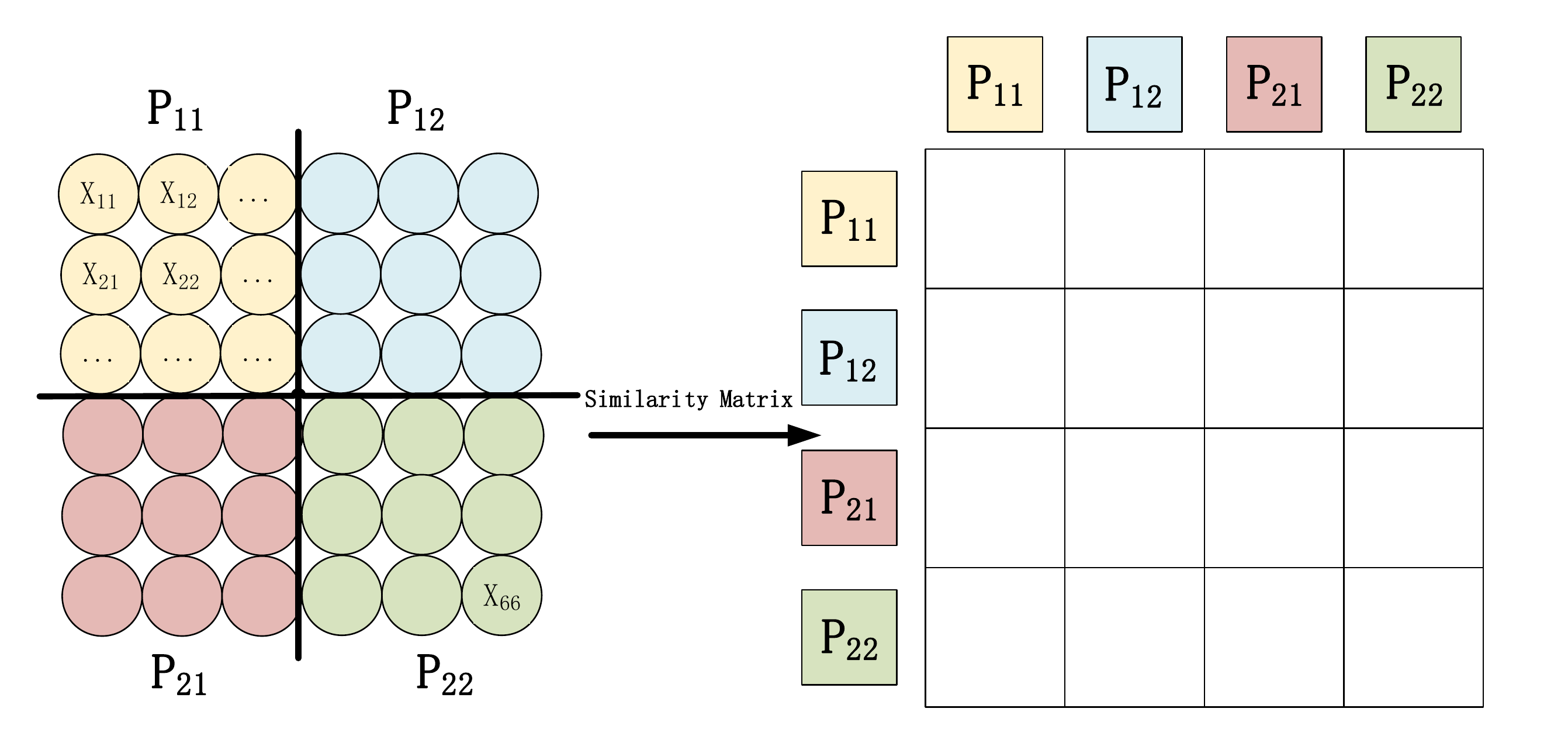}}
	\hspace{0.5in}
	\subfigure[The Similarity Matrix in terms of Region Attention.]{
		\label{f5:region_attention} 
		\includegraphics[width=0.7\linewidth]{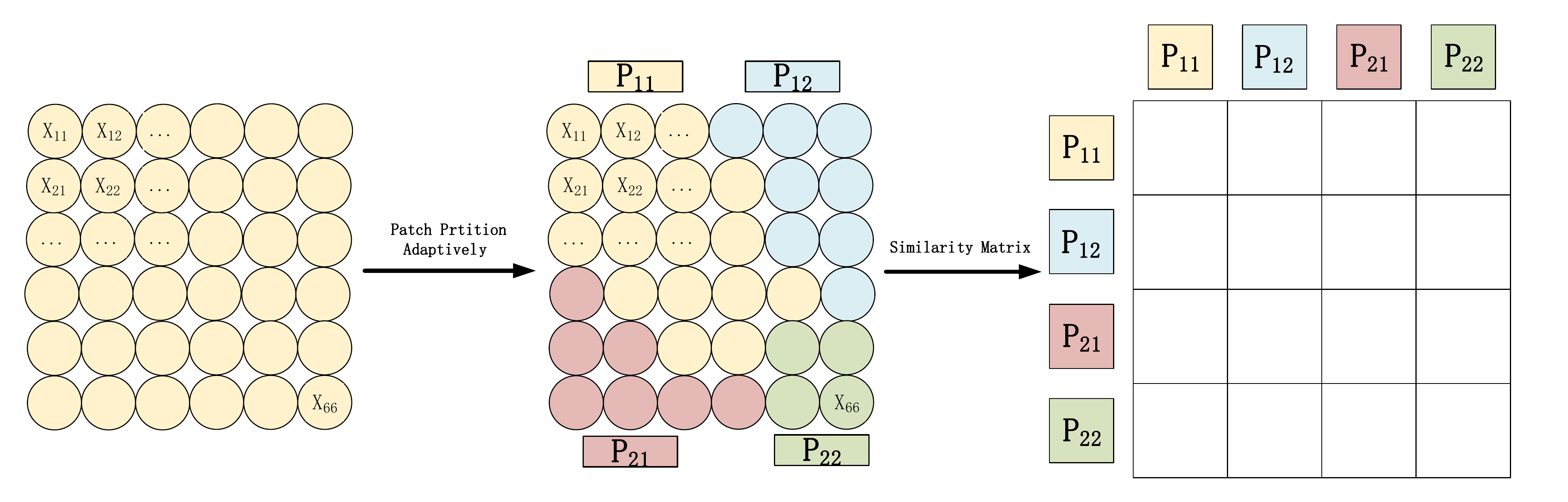}}
	\caption{The calculation process of Similarity Matrix based on three different attention methods}
	\label{fig:similarity_matrix} 
\end{figure*}

\subsection{Region Attention Blcok}
\label{Region Attention Blcok}

To capture the large-scale spatial feature maps, the spatial attention module is applied in the video prediction task \cite{lin2020self}. It calculates the similarity matrix by multiplying elements from different locations as Figure~\ref{f3:orgianl_attention} shown. However, over-fine grained operation between the elements often breaks the local representation. In order to preserve the whole local information, Vision Transformer (ViT) \cite{dosovitskiy2020image} directly divides the whole feature map into different patches and computes the similarity matrix between these regions in Figure~\ref{f4:vit_atetntion}. To this end, the local representation in different patches can interact with each other by attention operation. Therefore, it considers global and local information at the same time. However, the position of each patch is fixed in ViT model, which might cause the same object to be scattered in several neighbor patches and the local representation is destroyed to some extend.

\begin{figure*}[ht]
	\centering
	\includegraphics[width=1\linewidth]{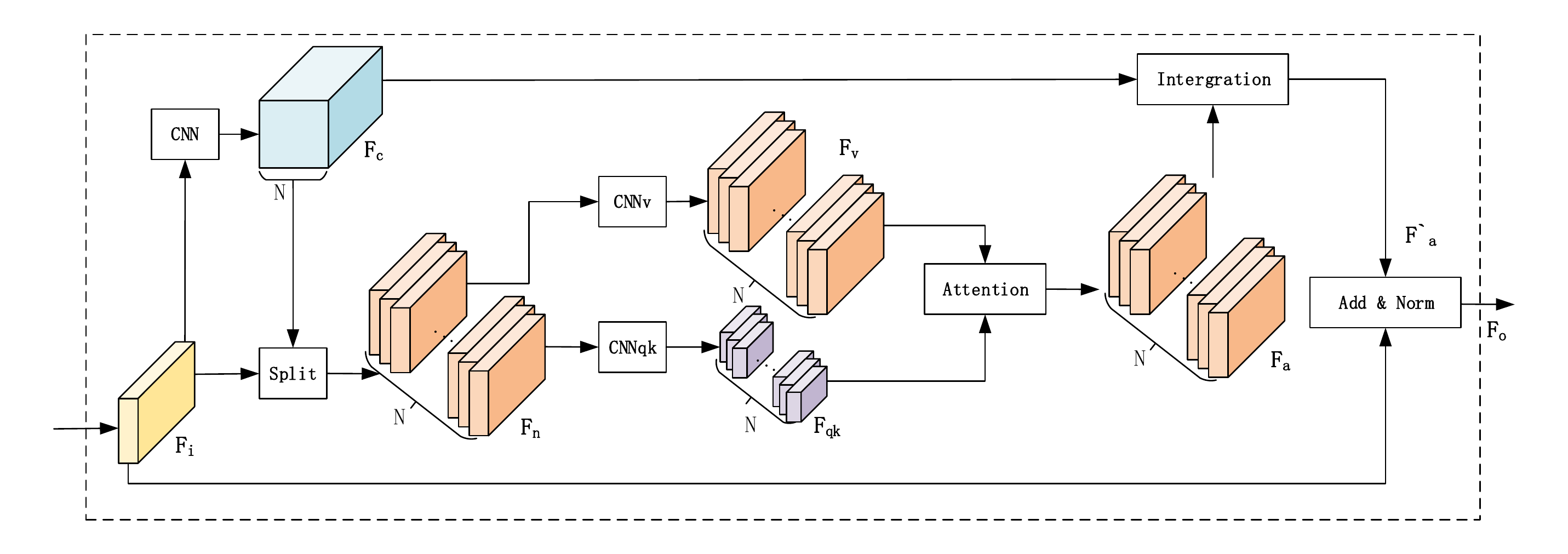}
	\caption{The structure of Region Attention Block (RAB).}
	\label{f6:region_attention_block}
\end{figure*}

To address the issue, we expect that these patches can be divided adaptively and those elements with similar semantic relationships are classified into the same patch shown in Figure~\ref{f5:region_attention}. To realize this idea, we propose the Region Attention Block (RAB) whose structure is illustrated in Figure~\ref{f6:region_attention_block}. 
First, a convolution and softmax layer are employed in the input feature map $F_i \in R^{B \times C \times H \times W}$ to generate $F_c \in R^{B \times N \times H \times W}$ for distinguishing $N$ classifications. Second, the orginal input $F_i$ is split $N$ groups of feature maps  $ F_b \in R^{N \times B \times C \times H \times W}$ by the Split Module following this formula:

\begin{equation}
\label{eq:split}
\begin{aligned}
\mbox{Split}(P, Q) = & \mbox{Concatenate}(\{P_{i,j,h,w} \cdot Q_{i,n,h,w}| 1<n<N, n\in Z\}, \\ 
&axis = 0), 
\end{aligned}
\end{equation}

\noindent where the $P$ and $Q$ indicate $F_i$ and $F_c$, respectively.  These groups denote various semantic information extract from different position. Third,  $F_{qk} \in R^{B \times N \times c \times h \times w}$ is convolved by $F_b$ to furthermore exploit the feature of $F_b$ and reduce the parameters, where the $c$, $h$ and $w$ are smaller than $C$, $H$ and $W$. Besides, $F_v \in R^{N \times B \times C \times H \times W}$ are outputed by a convolution layer applied in $F_b$ to preserve original information. Forth, three different convolutions are used to generate query $Q$, key $K$ and value $V$ based on $F_{qk}$ and $F_v$. After flattening, $Q \in R^{B \times N \times c*h*w}$, $K \in R^{B \times N \times c*h*w}$ and $V \in R^{B \times N \times C*H*W}$ are fed into the attention function to obtain $F_a \in R^{B \times N \times C \times H \times W}$ that have been interacted the local representation from different regions. Fifth,a Integration Module is utilized to consist $F_a$ based on the $F_c$ by this equation:

\begin{equation}
\label{eq:intergration}
\mbox{Integration(P, Q)} = \sum_{n=1}^{N} P_{i,j,h,w} \cdot Q_{i,n,j,h,w}.
\end{equation}

\noindent Here, $F_a$ and $F_c$ are represented by $P$ and $Q$, respectively. The result $F'_a$ has the same size of input feature map $F_i$. Finally, the structure of ResNet \cite{he2016deep} is introduced to achieve the final result $F_o \in R^{B \times N \times C \times H \times W}$. In summary, the calculation process is described by the following formulas:

\begin{small}
\begin{equation}
\begin{aligned}
\label{eq:region_attention}
F_c & = softmax(F_i * W_c),\\
F_n & = \mbox{Split}(F_i, F_c),\\
F_{qk} & = F_n*W'_{qk},\\
F_v & = F_n*W'_{v},\\
F_a & = \mbox{Attention}(F_{qk}*W_q,F_{qk}*W_k,F_v*W_v), \\
F'_a& = \mbox{Integration}(F_a,F_c), \\
F_o & = \mbox{Norm}(F_i+F_a).
\end{aligned}
\end{equation}
\end{small}

\subsection{Recall Attention Mechanism}

\label{Recall Attention Mechanism}

To capture the temporal long-dependencies of representation, Wang et.al \cite{wang2018eidetic} embedded the spatial attention module in the updating of temporal memory. However, it has two limitations 1): it saves abundant history temporal memories, which leads that the number of parameters easily exceeds the space occupancy as lead time goes. 2) The temporal memory has lost some information during the generation of various gates. Therefore, the preserved previous representation is not the whole information and long-term spatiotemporal expressivity is limited.

\begin{figure}[ht]
	\centering
	\includegraphics[width=0.75\linewidth]{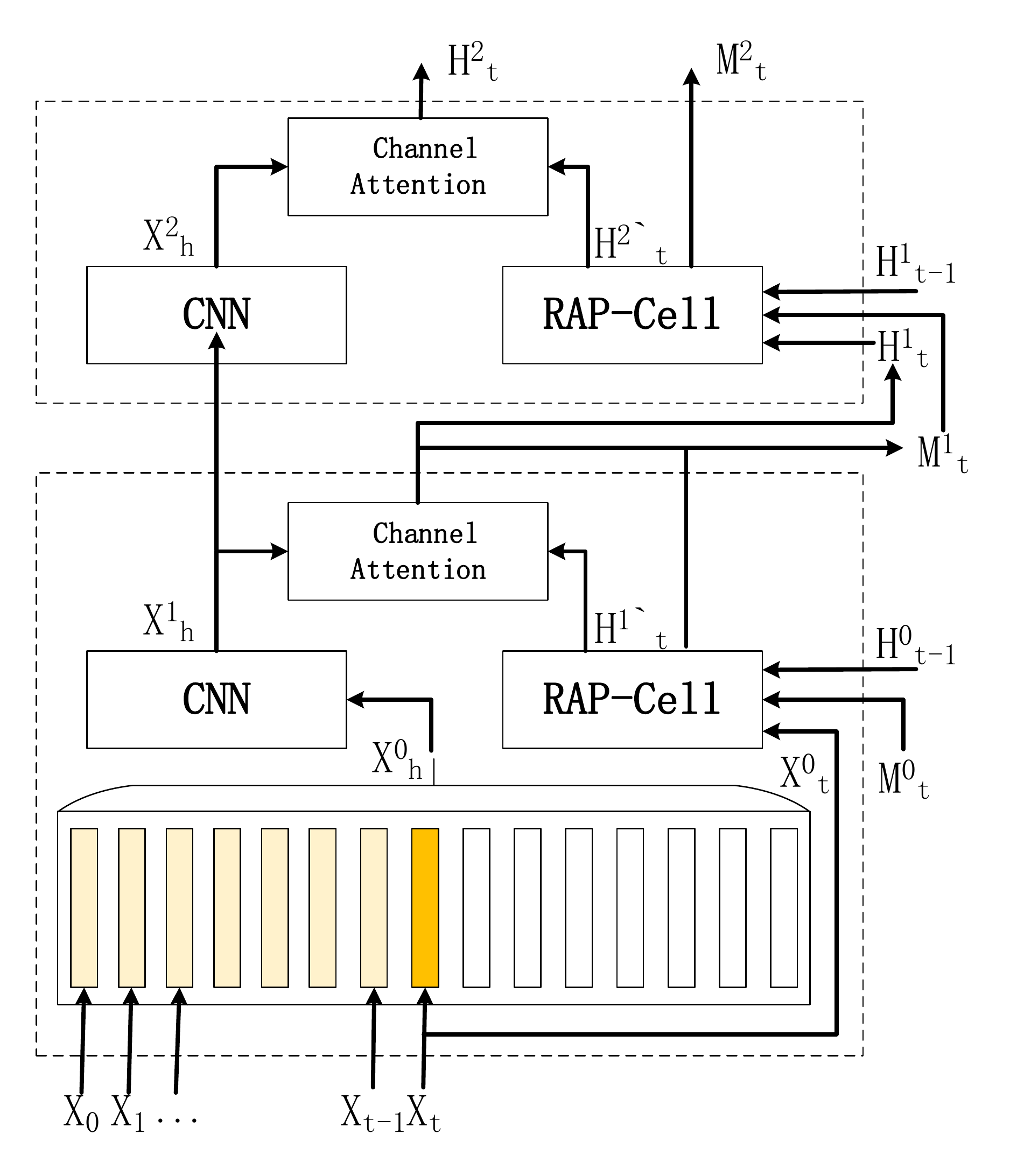}
	\caption{The structure of Recall Attention Mechanism (RAM).}
	\label{f7:recall_attention}
\end{figure}

To address these issues, we propose the Recall Attention Mechanism (RAM) to enhance the long-term spatiotemporal representation ability with fixed space occupation as Figure \ref{f7:recall_attention} shown. First, we build an empty long-memory feature map $X^0_h \in R^{B \times T \times C \times H \times W}$ in the bottom layer and feed into the current input $X_t$ continually. The  $X^0_h$ thus contains all original previous inputs $X_{0:t} \in R^{B \times C \times H \times W}$. Secondly, a layer convolution neural network is employed to extract the feature of $X^0_h$ and output the long-memory hidden state $X^1_h \in R^{B \times T \times C \times H \times W}$. Finally, $X^1_h$ and the output $H'_t$ of RAP-Cell feed into the channel attention module to generate new hidden states, where the $X^1_h$ can be regarded as the value and key, and the $H'_t$ represents the query. In this mechanism, the new output $H_t$ has recalled all original historical representation and long-term dependencies can be maximize preserved. Besides, the size of the long-memory feature map $X_h$ is limited at any time step. In the next layer, the input of the long-memory hidden state is the $X^1_h$ from the bottom layer. To this end, the long-term historical representation can deliver to the next layer.

\section{Experiments}
\label{Experimental}
\subsection{Dataset}

The dataset is collected from the CIKM AnalytiCup2017 competition and covers the whole area of Shenzhen. For convenience, we name this public dataset to RadarCIKM. RadarCIKM has a training set and test set with 10,000 and 4,000 sequences, respectively. There are 2,000 sequences randomly sampled from the training set to build the validation set.  Each sequence contains 15 continual observations within 90 minutes, where the spatial and temporal resolution of each map is 1km $\times$ 1km and six minutes, respectively. The range of each pixel is from 0 to 255 and each pixel $p$ can be converted to radar reflectivity (dBZ) by Z-R relationship as follows:  

\begin{equation}
\label{eq:ZR-relationship}
dBZ = p \times \frac{95}{255} -10.
\end{equation}


\subsection{Evaluation Metrics}

In this paper, except for common measurements such as Structural Similarity (SSIM), Mean Absolute Error(MAE) in video prediction, we also applied the Heidke Skill Score (HSS) and Critical Success Index (CSI) which are used in precipitation nowcasting task. The HSS and CSI evaluate the fraction of correct forecasts after eliminating random predictions and the number of correct forecasts divided by the total number of occasions when the rainfall events were forecasted or observed, respectively. Specifically, the prediction and ground truth are converted to binary matric based on a threshold $\tau$. Once the value of dBZ is larger than $\tau$, it is set to 1 otherwise to 0. Next, the number of the True-Positive (TP, prediction=1 and truth = 1), False-Negative (FN, prediction = 0 and truth=1), False-Positive (FP, prediction=1 and truth=0) and True-Negative (TN, prediction=0 and truth=0) are counted. Finally, the HSS and CSI can be calculated by the following formulas:

\begin{equation}
\label{eq:HSS}
dBZ =  \frac{2(TP \times TN - FN \times FP)}{(TP+FN)(FN+TN)+(TP+FP)(FP+TN)},
\end{equation}

\begin{equation}
\label{eq:CSI}
dBZ = \frac{TP}{TP+FN+FP}.
\end{equation}

\subsection{Parameters Setting}

The proposed RAP-Net takes five previous radar echo maps as inputs and outputs ten predictions. It utilizes four layers RAP-Units as shown in Figure \ref{f2:architecture}, where the number of patches is set to 64.  The Adam optimizer is applied to train our model with a 0.0001 learning rate. Besides, the early stopping and scheduled sampling strategies are also used to optimize our model. The loss function combines the L1 and L2 to train SST-LSTM. All experiments are implemented in Pytorch and conducted on NVIDIA 3090 Ti GPUs. The detail is available from the source code: \url{https://github.com/luochuyao/RAP-Net.}

\subsection{Result and Analysis}

Tables \ref{tab:hss_mae} and \ref{tab:csi_ssim} show the evaluations of all models.  We find that the RAP-Net achieves the smallest error and the highest structural similarity according to the MAE and SSIM. It shows that our model outperforms other models in terms of the whole prediction. Besides, the proposed model has significant superiority especially for the nowcasting in heavy rainfall regions. Because the HSS and CSI keep the top position in the middle and high thresholds (20 dBZ and 40 dBZ). For the state-of-art method, PFST-LSTM \cite{luo2020pfst}, all measurements of it are exceeded by RAP-Net, which shows the performance of our model furthermore. Comparing with PredRNN, PredRNN++ and RAP-Net, we can see that they have similar SSIM due to applying the same architecture. However, the other evaluate indexes of RAP-Net is the highest, which implies the advance of RAP-Unit. Finally, we notice the SA-ConvLSTM \cite{lin2020self} have the best HSS and CSI in the lowest threshold (5 dBZ) and bad performance in the highest performance, which means the Region Attention can improve the prediction in the area with high radar echo compared to traditional attention mechanism.

\begin{table}[ht]
	\centering
	\caption{Comparison results on RadarCIKM in terms of HSS and MAE}
	\label{tab:hss_mae}
	\setlength{\tabcolsep}{1mm}{
	\begin{tabular}{|c|cccc|c|}
		\hline
		\multirow{2}{*}{Methods} & \multicolumn{4}{c|}{HSS $\uparrow$}                                   & \multirow{2}{*}{MAE $\downarrow$} \\ \cline{2-5}
		& 5               & 20              & 40              & avg             &                                   \\ \hline
		ConvLSTM                 & 0.7035          & 0.4819          & 0.1081          & 0.4312          & 5.61                              \\
		ConvGRU                  & 0.6816          & 0.4827          & 0.1225          & 0.4289          & 6.00                              \\
		TrajGRU                  & 0.6809          & 0.4945          & 0.1907          & 0.4553          & 5.90                              \\
		DFN  \cite{jia2016dynamic}                    & 0.6772          & 0.4719          & 0.1306          & 0.4266          & 6.03                              \\
		PredRNN                  & 0.7081          & 0.4911          & 0.1559          & 0.4516          & 5.42                              \\
		PredRNN++                & 0.7075          & 0.4993          & 0.1575          & 0.4548          & 5.41                              \\
		MIM   \cite{wang2019memory}          		& 0.6959 		   & 
		0.4990 			& 0.1960 		  & 0.4636			& 5.65                    		  \\
		PhyDNet  \cite{guen2020disentangling}                & 0.6807          & 0.4725          & 0.1230          & 0.4254          & 5.99                              \\
		SA-ConvLSTM              & \textbf{0.7118} & 0.4861          & 0.1582          & 0.4520          & 5.82                              \\
		PFST-LSTM                & 0.7045          & 0.5071          & 0.2218          & 0.4778          & 5.82                              \\
		RAP-Net                  & 0.7117          & \textbf{0.5116} & \textbf{0.2293} & \textbf{0.4842} & \textbf{5.37}                     \\ \hline
	\end{tabular}
	}
\end{table}

\begin{table}[ht]
	\centering
	\caption{Comparison results on RadarCIKM in terms of CSI and SSIM}
	\label{tab:csi_ssim}
	\setlength{\tabcolsep}{1mm}{
	\begin{tabular}{|c|cccc|c|}
		\hline
		\multirow{2}{*}{Methods} & \multicolumn{4}{c|}{CSI $\uparrow$}                                   & \multirow{2}{*}{SSIM $\uparrow$} \\ \cline{2-5}
		& 5               & 20              & 40              & avg             &                                    \\ \hline
		ConvLSTM                 & 0.7656          & 0.4034          & 0.0578          & 0.4089          & 0.2118                             \\
		ConvGRU                  & 0.7522          & 0.3952          & 0.0657          & 0.4043          & 0.1360                             \\
		TrajGRU                  & 0.7466          & 0.4028          & 0.1061          & 0.4185          & 0.1532                             \\
		DFN  \cite{jia2016dynamic}                     & 0.7489          & 0.3771          & 0.0704          & 0.3988          & 0.1540                             \\
		PredRNN                  & 0.7691          & 0.4048          & 0.0854          & 0.4198          & 0.3100                             \\
		PredRNN++                & 0.7670          & 0.4137          & 0.0862          & 0.4223          & 0.3119                             \\
		MIM \cite{wang2019memory}            		 & 0.7530          & 
		0.3980          & 0.1095          & 0.4202			& 0.1697                    \\
		PhyDNet \cite{guen2020disentangling}                 & 0.7478          & 0.3882          & 0.0659          & 0.4006          & 0.1499                             \\
		SA-ConvLSTM              & \textbf{0.7725} & 0.4161          & 0.0870          & 0.4252          & 0.1548                             \\
		PFST-LSTM                & 0.7680          & 0.4175          & 0.1257          & 0.4371          & 0.1520                             \\
		RAP-Net                  & 0.7666          & \textbf{0.4305} & \textbf{0.1307} & \textbf{0.4426} & \textbf{0.3177}                    \\ \hline
	\end{tabular}
	}
\end{table}

\begin{figure}[ht]
	\centering
	\includegraphics[width=0.6\linewidth]{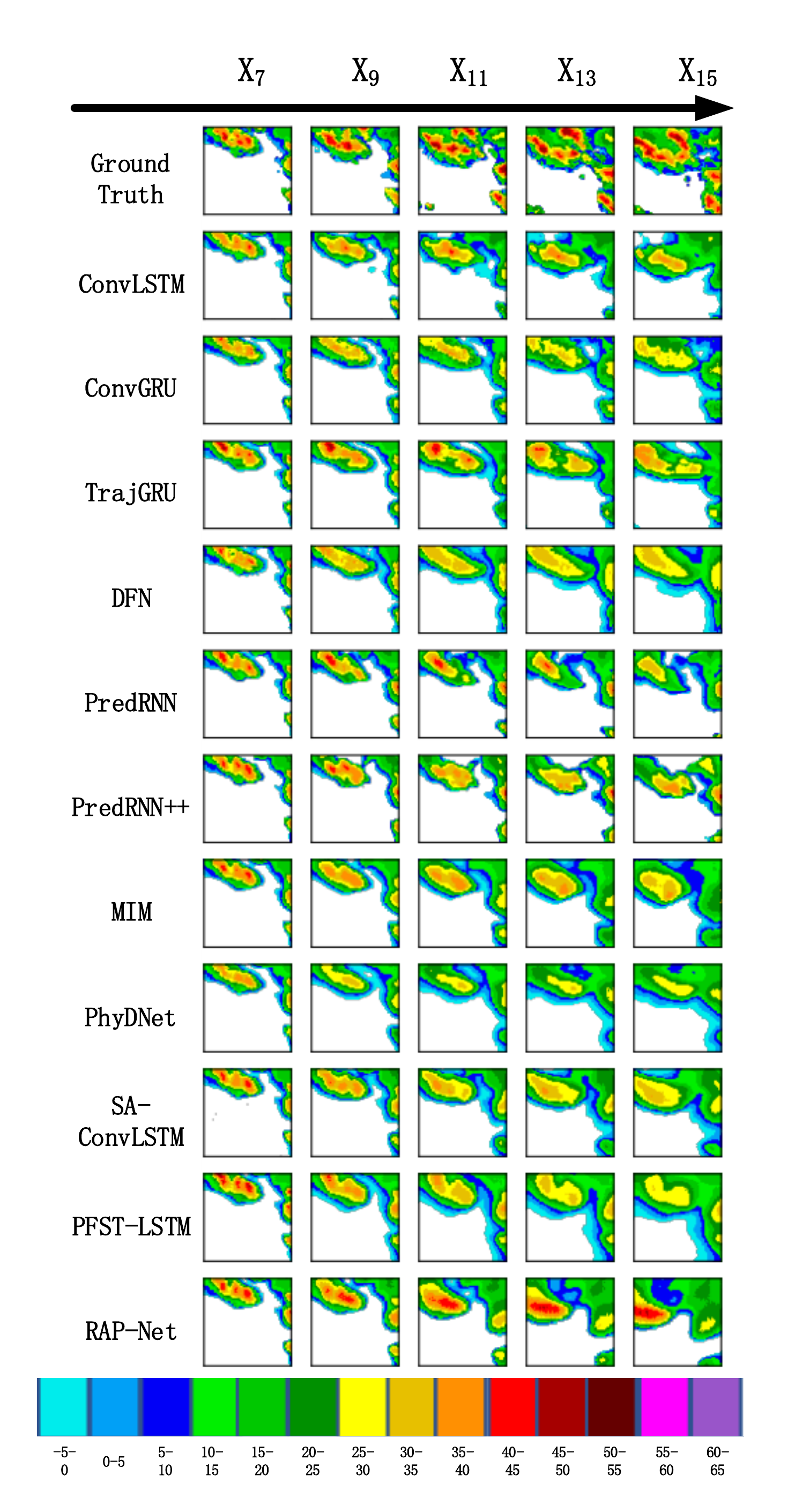}
	\caption{The first row is the ground truth and reminders are the predictions of all methods on an example from the RadarCIKM dataset (Best view in color)}
	\label{f8:res}	
\end{figure}

Figure~\ref{f8:res} shows an example of predictions from these models. The various colors denote the different ranges of reflectivity according to the color bar in the bottom of Figure~\ref{f8:res}. From the ground truth in the first row, the rainfall event is obviously the trend of increasing the rainfall intensity. However, only our model can forecast this trend and keep the intensity of the regions. The RAP-Net can generate a high reflectivity area, which can also explain why our model can achieve the highest evaluate index HSS and CSI in the middle and high thresholds. 

\subsection{Ablation Study}

To investigate the influence of various modules, we conduct an ablation study to discuss the function of Region Attention Block to the current input and the last hidden state. The result of evaluations is shown in Tables \ref{tab:ablation_hss_mae} and \ref{tab:ablation_csi_ssim}. RAP-Cell$_x$ and RAP-Cell$_h$ denote the PredRNN model embedding the RAB into the input and hidden state, respectively. RAP-Cell model is the combination of  RAP-Cell$_x$ and RAP-Cell$_h$, and can also be regarded as RAP-Net without RAM. The various measurements of RAP-Cell$_x$ and RAP-Cell$_h$ are higher than PredRNN, which shows the validation of introducing Region Attention Block. Besides, the HSS and CSI of RAP-Cell have significant improvements in the highest threshold, which implies that RAB simultaneously employed in the input and hidden state contributes to the prediction in the heavy rainfall regions. Moreover, By comparing the RAP-Cell and RAP-Net, we notice that the RAM can raise the accuracy of nowcasting especially in the area with middle-intensity rainfall.    

\begin{table}[ht]
	\centering
	\caption{Ablation results on RadarCIKM in terms of HSS and MAE}
	\label{tab:ablation_hss_mae}
	\setlength{\tabcolsep}{1mm}{
	\begin{tabular}{|c|cccc|c|}
		\hline
		\multirow{2}{*}{Methods} & \multicolumn{4}{c|}{HSS $\uparrow$}                                   & \multirow{2}{*}{MAE $\downarrow$} \\ \cline{2-5}
		& 5               & 20              & 40              & avg             &                                   \\ \hline
		PredRNN                  & 0.7081          & 0.4911          & 0.1558          & 0.4516          & 5.42                              \\
		RAP-Cell$_x$              & 0.7102          & 0.5042          & 0.1754          & 0.4633          & 5.36                             \\
		RAP-Cell$_h$              & 0.7149          & 0.4967          & 0.1753          & 0.4623          & \textbf{5.32}                              \\
		RAP-Cell                 & \textbf{0.7234} & 0.4757          & 0.2283          & 0.4758          & 5.64                              \\
		RAP-Net                  & 0.7117          & \textbf{0.5116} & \textbf{0.2293} & \textbf{0.4842} & 5.37                    
		    \\ 
		    \hline
	\end{tabular}
	}
\end{table}

\begin{table}[ht]
	\centering
	\caption{Ablation results on RadarCIKM in terms of CSI and SSIM}
	\label{tab:ablation_csi_ssim}
	\setlength{\tabcolsep}{1mm}{
	\begin{tabular}{|c|cccc|c|}
		\hline
		\multirow{2}{*}{Methods} & \multicolumn{4}{c|}{CSI $\uparrow$}                                   & \multirow{2}{*}{SSIM $\uparrow$} \\ \cline{2-5}
		& 5               & 20              & 40              & avg             &                                  \\ \hline
		PredRNN                  & 0.7691          & 0.4048          & 0.0854          & 0.4198          & 0.3100                           \\
		RAP-Cell$_x$              & 0.7747          & 0.4235          & 0.0967          & 0.4316          & 0.2979                           \\
		RAP-Cell$_h$              & 0.7772          & 0.4138          & 0.0967          & 0.4292          & 0.3065                           \\
		RAP-Cell                 & \textbf{0.7817} & 0.4143          & 0.1300          & 0.4420          & \textbf{0.3259}                  \\
		RAP-Net                  & 0.7666          & \textbf{0.4305} & \textbf{0.1307} & \textbf{0.4429} & 0.3177                           \\ \hline
	\end{tabular}
	}
\end{table}

\begin{figure}[ht]
	\centering
	\includegraphics[width=0.6\linewidth]{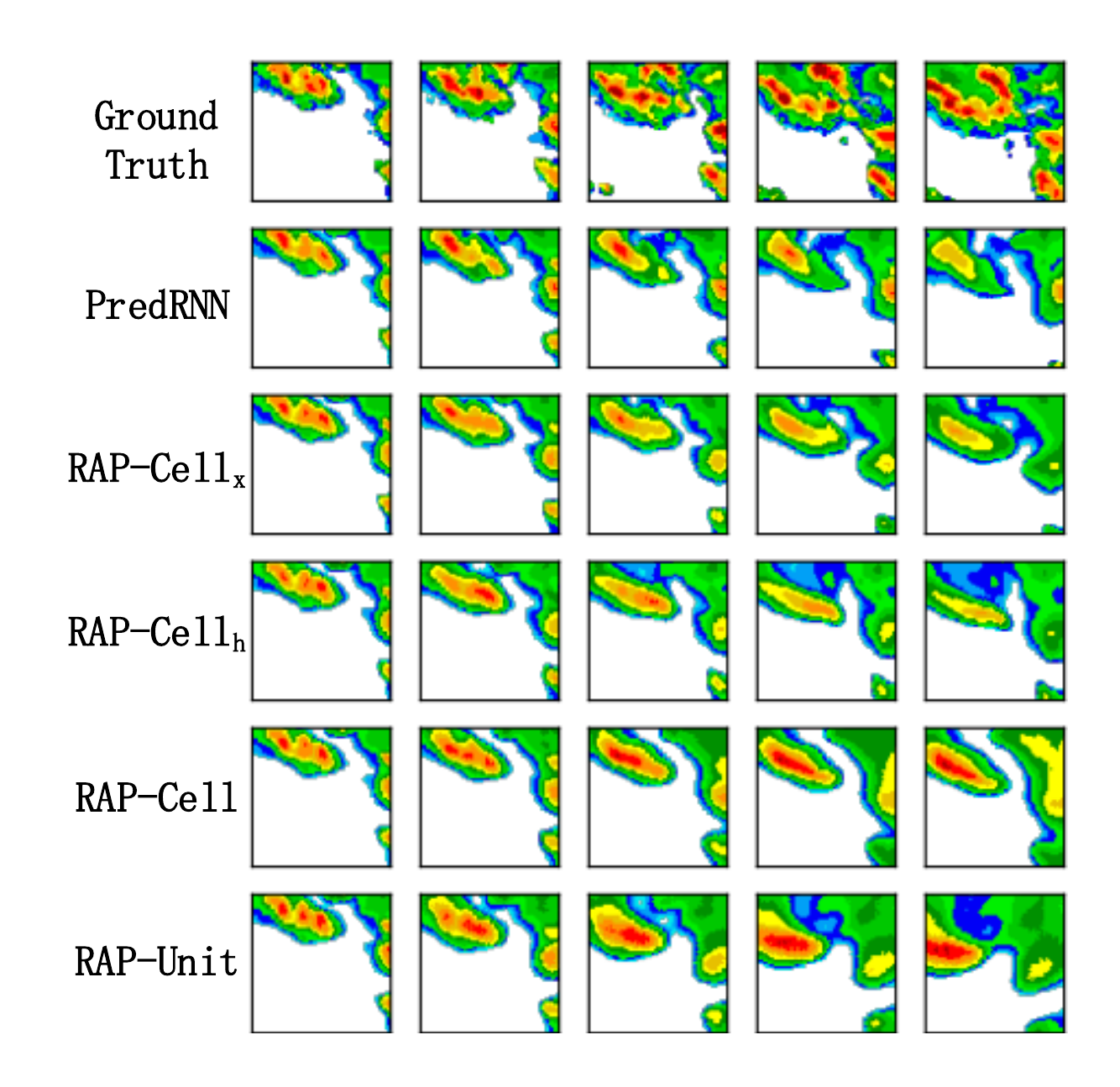}
	\caption{The first row is the ground truth and reminders are the predictions of all methods on an example from the RadarCIKM dataset (Best view in color)}
	\label{f9:ablation_res}	
\end{figure}

We also show predictions of the same sample in Figure \ref{f9:ablation_res}. we find that RAP-Cell can generate the red area which is reflected by better evaluation indexes of HSS and CSI in the highest threshold. However, all forecasts except for RAP-Net have a gap in the radar echo block, which is obviously different from the ground truth. The improvement of prediction in middle rainfall intensity can be owed to the embedding of RAM.

\section{Conclusion}

In this paper, we propose the RAP-Net to handle the precipitation nowcasting task. On the one hand, It embeds the Region Attention Block to enhance the local and global spatial representation ability simultaneously as extracting and delivering the features in ConvRNN. The improvement significantly enhances the accuracy especially in those regions with heavy rainfall. On the other hand, we introduce the Recall Attention Mechanism to improve the temporal expressivity in the long term. It can preserve and retrieve longer historical information and effectively enhance the performance of prediction, particularly for the middle rainfall intensity. In the future, we will explore how to predict longer radar echo maps.

\bibliographystyle{unsrtnat}
\bibliography{references}

\begin{thebibliography}{18}
\providecommand{\natexlab}[1]{#1}
\providecommand{\url}[1]{\texttt{#1}}
\expandafter\ifx\csname urlstyle\endcsname\relax
  \providecommand{\doi}[1]{doi: #1}\else
  \providecommand{\doi}{doi: \begingroup \urlstyle{rm}\Url}\fi

\bibitem[Shi et~al.(2017)Shi, Gao, Lausen, Wang, Yeung, Wong, and
  Woo]{shi2017deep}
Xingjian Shi, Zhihan Gao, Leonard Lausen, Hao Wang, Dit-Yan Yeung, Wai-kin
  Wong, and Wang-chun Woo.
\newblock Deep learning for precipitation nowcasting: A benchmark and a new
  model.
\newblock \emph{arXiv preprint arXiv:1706.03458}, 2017.

\bibitem[Woo and Wong(2017)]{woo2017operational}
Wang-chun Woo and Wai-kin Wong.
\newblock Operational application of optical flow techniques to radar-based
  rainfall nowcasting.
\newblock \emph{Atmosphere}, 8\penalty0 (3):\penalty0 48, 2017.

\bibitem[Xingjian et~al.(2015)Xingjian, Chen, Wang, Yeung, Wong, and
  Woo]{xingjian2015convolutional}
SHI Xingjian, Zhourong Chen, Hao Wang, Dit-Yan Yeung, Wai-Kin Wong, and
  Wang-chun Woo.
\newblock Convolutional lstm network: A machine learning approach for
  precipitation nowcasting.
\newblock In \emph{Advances in neural information processing systems}, pages
  802--810, 2015.

\bibitem[Wu et~al.(2021)Wu, Yao, Wang, and Long]{wu2021motionrnn}
Haixu Wu, Zhiyu Yao, Jianmin Wang, and Mingsheng Long.
\newblock Motionrnn: A flexible model for video prediction with
  spacetime-varying motions.
\newblock In \emph{Proceedings of the IEEE/CVF Conference on Computer Vision
  and Pattern Recognition}, pages 15435--15444, 2021.

\bibitem[Lin et~al.(2020)Lin, Li, Zheng, Cheng, and Yuan]{lin2020self}
Zhihui Lin, Maomao Li, Zhuobin Zheng, Yangyang Cheng, and Chun Yuan.
\newblock Self-attention convlstm for spatiotemporal prediction.
\newblock In \emph{Proceedings of the AAAI Conference on Artificial
  Intelligence}, volume~34, pages 11531--11538, 2020.

\bibitem[Wang et~al.(2018{\natexlab{a}})Wang, Jiang, Yang, Li, Long, and
  Fei-Fei]{wang2018eidetic}
Yunbo Wang, Lu~Jiang, Ming-Hsuan Yang, Li-Jia Li, Mingsheng Long, and
  Li~Fei-Fei.
\newblock Eidetic 3d lstm: A model for video prediction and beyond.
\newblock In \emph{International conference on learning representations},
  2018{\natexlab{a}}.

\bibitem[Wang et~al.(2013)Wang, Wong, Liu, and Wang]{wang2013application}
Gaili Wang, Waikin Wong, Liping Liu, and Hongyan Wang.
\newblock Application of multi-scale tracking radar echoes scheme in
  quantitative precipitation nowcasting.
\newblock \emph{Advances in Atmospheric Sciences}, 30\penalty0 (2):\penalty0
  448--460, 2013.

\bibitem[Ryu et~al.(2020)Ryu, Lyu, Do, and Lee]{ryu2020improved}
Soorok Ryu, Geunsu Lyu, Younghae Do, and GyuWon Lee.
\newblock Improved rainfall nowcasting using burgers’ equation.
\newblock \emph{Journal of Hydrology}, 581:\penalty0 124140, 2020.

\bibitem[Wang et~al.(2017)Wang, Long, Wang, Gao, and Yu]{wang2017predrnn}
Yunbo Wang, Mingsheng Long, Jianmin Wang, Zhifeng Gao, and Philip~S Yu.
\newblock Predrnn: Recurrent neural networks for predictive learning using
  spatiotemporal lstms.
\newblock In \emph{Proceedings of the 31st International Conference on Neural
  Information Processing Systems}, pages 879--888, 2017.

\bibitem[Wang et~al.(2018{\natexlab{b}})Wang, Gao, Long, Wang, and
  Philip]{wang2018predrnn++}
Yunbo Wang, Zhifeng Gao, Mingsheng Long, Jianmin Wang, and S~Yu Philip.
\newblock Predrnn++: Towards a resolution of the deep-in-time dilemma in
  spatiotemporal predictive learning.
\newblock In \emph{International Conference on Machine Learning}, pages
  5123--5132. PMLR, 2018{\natexlab{b}}.

\bibitem[Shouno(2020)]{shouno2020photo}
Osamu Shouno.
\newblock Photo-realistic video prediction on natural videos of largely
  changing frames.
\newblock \emph{arXiv preprint arXiv:2003.08635}, 2020.

\bibitem[Xie et~al.(2020)Xie, Li, Ji, Chen, Chen, Liu, and Ye]{xie2020energy}
Pengfei Xie, Xutao Li, Xiyang Ji, Xunlai Chen, Yuanzhao Chen, Jia Liu, and
  Yunming Ye.
\newblock An energy-based generative adversarial forecaster for radar echo map
  extrapolation.
\newblock \emph{IEEE Geoscience and Remote Sensing Letters}, 2020.

\bibitem[Dosovitskiy et~al.(2020)Dosovitskiy, Beyer, Kolesnikov, Weissenborn,
  Zhai, Unterthiner, Dehghani, Minderer, Heigold, Gelly,
  et~al.]{dosovitskiy2020image}
Alexey Dosovitskiy, Lucas Beyer, Alexander Kolesnikov, Dirk Weissenborn,
  Xiaohua Zhai, Thomas Unterthiner, Mostafa Dehghani, Matthias Minderer, Georg
  Heigold, Sylvain Gelly, et~al.
\newblock An image is worth 16x16 words: Transformers for image recognition at
  scale.
\newblock In \emph{International Conference on Learning Representations}, 2020.

\bibitem[He et~al.(2016)He, Zhang, Ren, and Sun]{he2016deep}
Kaiming He, Xiangyu Zhang, Shaoqing Ren, and Jian Sun.
\newblock Deep residual learning for image recognition.
\newblock In \emph{Proceedings of the IEEE conference on computer vision and
  pattern recognition}, pages 770--778, 2016.

\bibitem[Luo et~al.(2020)Luo, Li, and Ye]{luo2020pfst}
Chuyao Luo, Xutao Li, and Yunming Ye.
\newblock Pfst-lstm: A spatiotemporal lstm model with pseudoflow prediction for
  precipitation nowcasting.
\newblock \emph{IEEE Journal of Selected Topics in Applied Earth Observations
  and Remote Sensing}, 14:\penalty0 843--857, 2020.

\bibitem[Jia et~al.(2016)Jia, De~Brabandere, Tuytelaars, and
  Gool]{jia2016dynamic}
Xu~Jia, Bert De~Brabandere, Tinne Tuytelaars, and Luc~V Gool.
\newblock Dynamic filter networks.
\newblock \emph{Advances in neural information processing systems},
  29:\penalty0 667--675, 2016.

\bibitem[Wang et~al.(2019)Wang, Zhang, Zhu, Long, Wang, and Yu]{wang2019memory}
Yunbo Wang, Jianjin Zhang, Hongyu Zhu, Mingsheng Long, Jianmin Wang, and
  Philip~S Yu.
\newblock Memory in memory: A predictive neural network for learning
  higher-order non-stationarity from spatiotemporal dynamics.
\newblock In \emph{Proceedings of the IEEE/CVF Conference on Computer Vision
  and Pattern Recognition}, pages 9154--9162, 2019.

\bibitem[Guen and Thome(2020)]{guen2020disentangling}
Vincent~Le Guen and Nicolas Thome.
\newblock Disentangling physical dynamics from unknown factors for unsupervised
  video prediction.
\newblock In \emph{Proceedings of the IEEE/CVF Conference on Computer Vision
  and Pattern Recognition}, pages 11474--11484, 2020.

\end{thebibliography}

\end{document}